# IPA Transcription of Bengali Texts

**Kanij Fatema**[1], **Fazle Dawood Haider**[2], **Nirzona Ferdousi Turpa**[2], **Tanveer Azmal**[2,1], **Sourav Ahmed**[1,3]
**Navid Hasan**[1,3], **Mohammad Akhlaqur Rahman**[2,3], **Biplab Kumar Sarkar**[3,5], **Afrar Jahin**[3,5]
**Md. Rezuwan Hassan**,[3,4] **Md Foriduzzaman Zihad**,[4,1] , **Rubayet Sabbir Faruque**,[4,1] , **Asif Sushmit**[1,*]
**Mashrur Imtiaz**[2], **Farig Sadeque**[4], **Syed Shahrier Rahman**[2]

[1]Bengali.AI   [2]University of Dhaka
[3]Shahjalal University of Science and Technology   [4]BRAC University   [5]Sylhet Engineering College

**Abstract**

The International Phonetic Alphabet (IPA) serves to systematize phonemes in language, enabling precise textual representation of pronunciation. In Bengali phonology and phonetics, ongoing scholarly deliberations persist concerning the IPA standard and core Bengali phonemes. This work examines prior research, identifies current and potential issues, and suggests a framework for a Bengali IPA standard, facilitating linguistic analysis and NLP resource creation and downstream technology development. In this work, we present a comprehensive study of Bengali IPA transcription and introduce a novel IPA transcription framework incorporating a novel dataset with DL-based benchmarks.

## 1. Introduction

Bangla, popularly known as Bengali, is the official language of Bangladesh and is spoken by a vast population of 272.7 million people in Bangladesh and some regions of India, along with a massive Bengali diaspora all across the globe. In various communities, Bangla-speaking people use a wide variety of dialects of the language. The morphological variations among these dialects are relatively subtle, but distinctions are found in sounds and phonology. This calls for a consistent IPA transcription protocol for canonical and dialectal variations of Bangla, to document the language well.

### 1.1. International Phonetic Alphabet (IPA)

A Bangla-to-IPA transcription model requires a phonetic transcription scheme to represent the transcription and pronunciation patterns for the language. The International Phonetic Alphabet (IPA) stands as the sole standard for phonetic writing systems.Regardless of the language in question, the International Phonetic script relies on Roman characters as well as incorporates modified elements from diverse scripts like Greek to convey phonetic notation. The IPA-provided symbols such as (t, ɛ, ʃ, k, ̪ ) are to be used for even those language that does not employ the Roman alphabet, such as Bangla, Hindi, Japanese, or Korean.

Since its establishment in 1886, the International Phonetic Association has concerned itself with developing a system of symbols that maintains a balance between usability and inclusivity, which encompasses the wide variety of sounds present in languages all over the world (Association, 1999). The main purpose of IPA is to represent specific speech sounds rather than the abstract linguistic units known as phonemes, although it is also used for phonemic transcription (Association, 1999). IPA follows a common policy of using one letter for each segment. As a result, two letters are not put together to represent one single sound. For example, In the word 'shine', 'sh' is used to convey one single sound. IPA doesn't usually provide separate characters for sounds that aren't differentiated in known languages. Both broad and narrow transcriptions can be used in IPA. Details on IPA representation of Vowels, Consonants, Suprasegmentals and Diacritics are shown in Appendix section.

### 1.2. Literature Review: Bangla IPA

Bangla (internationally popular term *Bengali* is used interchangeably in this paper as well) possesses a distinctive phonetic inventory, which can be represented using the IPA. The requirement of an IPA transcription model is a phonetic transcription scheme to represent the pronunciation patterns for the language. Numerous studies have delved into the standard IPA representation for Bangla, developing a range of perspectives and viewpoints. Recently, a government-sanctioned IPA website has been introduced in Bangladesh. This platform adheres to the standard International Phonetic Alphabet (IPA) corresponding to the Bangla language, as outlined in the revised 2015 version. It encompasses seven vowels ই/i/, এ/e/, অ্যা/æ/, আ/a/, অ/ɔ/, ও/o/, and উ/u/ and provides suggestions for semi-vowels, specifying their corresponding IPA symbols as following: ই শ্রুতি/j/, অ/য় শ্রুতি/y/ and ব শ্রুতি/w/. Although, alternative representations are proposed for specific cases, such as ও/o̞/, ই/i̞/ এ/e̞/ and ঊ/u̞/. In terms of consonants, the website employs a set of 31 phonemes. For voiced consonants, they have provided both voiced /ʰ/ and voiceless/ʱ/ aspiration such as for ড /dʰ/ and /dʱ/, for ধ /d̪ʰ/ and /d̪ʱ/, for ভ /bʰ/ and /bʱ/, for ঢ /ɽʰ/ and /ɽʱ/. For য়, they provided the /y/, even though this sound does not exist in the Bangla language.

---

[*]Corresponding author: sushmit@ieee.org, farig.sadeque@bracu.ac.bd, mashrur.imtiaz@du.ac.bd, ss.rahman@du.ac.bd

This section further delves into these discussions before exploring the suggested IPA protocol for this dataset and outlining the validation challenges.

### 1.2.1. Bangla Vowels

Chatterji (1921) used Jones (1922)'s cardinal vowel system to explain the Bangla vowel system. He claimed that the Bangla language has seven primary vowels ই/i/, এ/e/, অ্যা/æ/, আ/a/, ও/o/, অ/ɔ/, and ঊ/u along with their corresponding nasal counterparts /ĩ ẽ æ̃ ã õ ɔ̃ ũ/. Chatterji also noted that Bangla vowels are generally articulated in a lax manner, imparting the characteristic 'timbre' to the vowel system. Morshed (1997) categorized the vowels as /i, u, e, o, ae, ɔ, and a/, including two high, two high-mid, two low-mid, and one low vowel. Ali (2001) investigated vowel contrasts, defining phonological properties, and reported the same number of vowels, with a subtle distinction. He employed the symbol /ɛ/ to represent the vowel /æ/ as described by Morshed (1997).

In a separate study, Hai (1964) analyzed the vowels of Standard Bangla using the concept of cardinal vowels. He claimed that there are eight vowels ই/i/, এ/e/, অ্যা/æ/, আ/a/a, ও/o/ ও'/o'/, অ/ɔ/, and ঊ/u/ in the Bangla language. He categorizes ই/i/, এ/e/, অ্যা/æ/ as front vowel and ও/o/, ও'/o'/, অ/ɔ/, and ঊ/u/ as back vowel. In contrast to Morshed (1997), Hai did not classify the Bangla vowel আ/a/a as occupying a central position. He explained that the Bangla আ/a/ sound differs from the neutral quality of the English /a/ and is distinct from the Urdu close /ə/ sound. Instead, he characterized it as an open vowel. Hai also pointed out the presence of an additional vowel in the Bangla vowel system, denoted as /o'/. He explained that when producing the /o'/ sound, the lips are slightly less rounded compared to the /o/ sound. However, there isn't a significant difference in the gap between the jaws, and the back of the tongue is not raised as much as it is when articulating the /o/ sound. This led him to term it as yotized o (oʸ), known in Bangla as অভিশ্রুত /obʱɪsɾut̪o/ ও /o/ or ও' /o'/. This observation was supported by Huq (2002). An example provided for this distinction is between বিয়ের ক'নে/bɪʲeɾ ko'ne/ and ঘরের কোণে /gʱɔɾeɾ kone/. Nevertheless, it's worth noting that there is limited empirical evidence to support this concept. On the contrary, the claim that the number of vowels is seven is backed by Pobitro Sorkar (1992) and Puny Sloka Ray (1997) as noted in Ali (2001).

### 1.2.2. Bangla Semi-Vowels

According to Chatterji (1921) and Sen (1993), there are two Bangla semivowels, namely অন্তস্থ ব/w/ and অন্তস্থ য় /y/. Hai (1964) contends that there are three semivowels: অন্তস্থ ব/w/, অন্তস্থ য় /y/, and অন্তস্থ ই/i/. Morshed (1997) argues that while অন্তস্থ ব/w/ and অন্তস্থ য় /y/ are considered semivowels in English, they do not possess similar status in Bangla. A different perspective was presented by Ferguson and Chowdhury (1960), who claim that there are four semivowels: /i e o u/. It is noted in Ali (2001) that this assertion was supported by Pobitro Sharker and Ghonesh Boshu (1998). Along with the ই/i̞/, ঊ/u̞/, and ও/o̞/, there is a fourth semi-vowel which is এ/e̞/ that is found at the end of the word in the form of 'য়' such as হয়/hɔe̞/, যায়/jae̞/ (Ali, 2001).

### 1.2.3. Bangla Diphthongs

Sen (1993) noted that the Bangla has two diphthongs: ঐ(oɪ) and ঔ(ou). These combinations of two sounds do not fit the conventional definition of diphthongs but are represented in written form. In linguistic terms, they are referred to as digraphs (Ali, 2001). On the contrary, Chatterji (1921) claimed that there are 25 diphthongs in standard Bangla. Hai (1964) asserted that there are a total of 31 diphthongs, categorizing them into 19 regular and 12 irregular ones. However, he also once argued that there are only 18 diphthongs, as noted by Ali (2001), who in turn asserts that there are 17 diphthongs in Bangla. The government-approved IPA website acknowledges the regular 19 diphthongs, but they have used the diphthong /ui̞/ two times and did not consider the /eo̞/ diphthong.

### 1.2.4. Bangla Consonants

There have been numerous past studies, primarily rooted in articulatory phonetics, that have examined the articulatory and acoustic characteristics of Bangla consonants. It is described in (Hai, 1964) that Bangla consonant has 20 stops, 7 fricatives, 4 nasals, 1 lateral, 1 trill, 2 flaps, and 1 glide; totaling 36 consonants. Hai (1964) claims that there's only on phone close to /ʃ/ in Bangla. Huq (2002) presented a slightly different categorization of a total of 35 consonants, presenting 21 stops, 5 fricatives, 3 nasals, 1 lateral, 1 trill, 2 flaps, and 2 glides. Morshed (1997) stated that Bangla includes 20 stops, 4 nasals, 4 fricatives, 1 lateral, and 2 flaps, totaling 31 consonants. On the other hand, Ali (2001) argued that Bangla has 20 stops, 3 nasals, 3 fricatives, 1 lateral, 2 flaps, 1 trill, and 2 glides, resulting in a total of 32 consonants.

### 1.3. Our Contribution

In this work, we present A **comprehensive study** of IPA transcription issues and challenges for Bangla, a novel **IPA transcription framework**, a **DUAL-IPA**, a sentence level ipa transcribed paral-lel corpus of 150k samples and DL-based benchmarking results. We open-source the dataset with the CC BY-SA 4.0 license.

## 2. Bangla IPA Transcription

Despite the global use of the Bangla language, there's a notable absence of a comprehensive IPA transcription framework and modeling. While the government-endorsed IPA system exists, it doesn't always offer clear explanations for specific diacritic usage, nor does it provide consistent reasoning for transcribing loaned words, accounting for morphological variations, or giving accurate IPA transcriptions. Besides, there remain unresolved debates among linguists regarding the inventory of vowels, semi-vowels, diphthongs, and consonants in Bangla. Scholars like Hai (1964) have observed that the existence of long vowels in the language does not make a difference in the meaning and specific tongue positions for vowel/a/, which leads us to questions about the articulation manner of morphological suffixes and accurate numbers of pure vowels in the language.

Regional variations of the Bangla language further complicate matters, impacting not only the pronunciation variation among individual speakers but also how sounds are produced based on different regions and dialects. Noting all these drawbacks of the Bangla language, we propose an IPA framework that we've employed to create a dataset of 70,000 words, alongside a modeling approach for accurate Bangla-to-IPA transcription. It's worth mentioning that our suggested phonetic representations may not be universally accepted, and users are encouraged to substitute specific phonemes with alternatives that better align with their linguistic preferences. With the readily available IPA chart, individuals can easily determine which sounds best match the intended IPA representation.

### 2.1. Vowels

In our proposed IPA, we conducted a thorough review and made some revisions that were then incorporated into our dataset. It's important to note that the vowel sounds in Bangla are articulated in a lax manner. After carefully listening to the IPA sounds provided by Ladefoged and Johnson (2014), we devised a chart where we recommend substituting /ɐ/ for /a/ when representing the Bangla letter 'আ'. The /a/ is an open vowel and it's produced towards the front of the mouth. On the other hand, /ɐ/ is produced at the center of the mouth and the mouth is slightly less open while articulating this which is more suitable for the Bangla letter 'আ' rather than the /a/ sound. Similarly, for the Bangla letter 'ই', we propose representing it as /ɪ/. The position of /ɪ/ is a near-high, front vowel in comparison to /i/ which is a high, front vowel. While producing the /ɪ/ sound, the position of the tongue remains slightly lower and back in the mouth in comparison to the /i/. The reason we propose /ɪ/ for the Bangla letter 'ই' is that the /ɪ/is a lax vowel and when we produce the 'ই' sound, there is less muscular tension in the tongue. This adjustment better aligns with the articulation of native Bangla speakers, where the /ɐ/ and /ɪ/ sounds are more appropriate. Regarding the আা sound, both /æ/ and /ɛ/ are true equivalents. However, for consistency in our dataset, we have chosen to use /ɛ/ exclusively.

|          | Front | Central | Back |
|----------|-------|---------|------|
| High     | ɪ     |         | u    |
| High-mid | e     |         | o    |
| Low-mid  | æ/ɛ   |         | ɔ    |
| Low      |       | ɐ       |      |

Table 1: Bangla Proposed Vowel Chart

### 2.2. Semi-vowel

Semi-vowels, often referred to as glides or semi-consonants, are phonetically identical to vowels but function as the syllable's boundary rather than as the nucleus, which is the central component of the syllable. In the International Phonetic Alphabet (IPA), the arch diacritic ( ̯ ) which is an inverted breve is used beneath semi-vowels to denote their dual nature, exhibiting features of both vowels and consonants. We have proposed four semi-vowels that have been incorporated into the dataset.

Those are given below in **( Bangla, /IPA/)** template, (ই, /ɪ̯/), (উ, /u̯/), (ও, /o̯/) and (এ, /e̯/)

### 2.3. Diphthongs

Hai (1964) provided a list of 31 Bangla diphthongs among which 19 diphthong (ɐɪ̯, ɐe̯, ɐu̯, ɐo̯, ɛe̯, ɛo̯, ɔe̯, ɔo̯, eɪ̯, eu̯, oɪ̯, oe̯, ou̯, oo̯, ɪɪ̯, ɪu̯, uɪ̯, uu̯, eo̯) are commonly found in the Bangla language. He further explores the Bangla diphthongs and claims that there are extra 12 diphthongs (ɪe̯, ɪa, ɪo̯, ea, eo̯, æa, oa, oe, ue, ua, uo) occurs irregularly.

To maintain clarity, it's wise to include all 31 diphthongs, especially considering the presence of regional dialects that might feature words absent in standard Bangla. Moreover, accurately discerning diphthongs requires audio reference rather than relying solely on written text. It's essential to acknowledge irregular diphthongs, particularly those involving the /a/ sound, which lacks a semi-vowel counterpart in Bangla. Therefore, the determination of whether a diphthong is rising or falling as well as whether is a vowel cluster or actually a diphthong hinges on careful consideration.

| Place \ Manner | | Bilabial | | Dental | | Alveolar | | Post-Alveolar | Palatal | | Velar | | Glottal |
|---|---|---|---|---|---|---|---|---|---|---|---|---|---|
| | | Unasp | Asp | Unasp | Asp | Unasp | Asp | | Unasp | Asp | Unasp | Asp | |
| Stop | Voiceless | প/p/ | ফ/pʰ/ | ত/t̪/ | থ/t̪ʰ/ | ট/t/ | ঠ/tʰ/ | | চ/c/ | ছ/cʰ/ | ক/k/ | খ/kʰ/ | |
| | Voiced | ব/b/ | ভ/bʱ/ | দ/d̪/ | ধ/d̪ʱ/ | ড/d/ | ঢ/dʱ/ | | জ, য /ɟ/ | ঝ/ɟʱ/ | গ/g/ | ঘ/gʱ/ | |
| Nasal | | ম/m/ | | | | ন, ণ/n/ | | | | | ঙ, ○ং/ŋ/ | | |
| Tap | | | | | | র /ɾ/ | | | | | | | |
| Flap | | | | | | ড়/ɽ/, ঢ়/ɽʰ/ | | | | | | | |
| Fricatives | | | | | | শ, স/s/ | | শ, ষ, স/ʃ/ | | | | | *হ/h/ |
| Lateral | | | | | | ল/l/ | | | | | | | |
| Approximant | | | | | | | | | *য/j/ | | | | |

Table 2: Proposed Consonant Chart. Here, Unasp. is used for unaspirated, and Asp. is used for aspirated

## 2.4. Consonants

In certain contexts, the 'হ' /h/ have extra careful articulation. For example, the word 'হ্রাস' in normal conversation would be pronounced as /ɾɐʃ/ but a news presenter or a person reciting a poem would articulate with an aspiration sound in the initial position of the word such as /ʰɾɐʃ/, following a more accepted canonical standard.

In the Bangla language, the য় /j/ is not articulated as a phoneme but is commonly used in the co-articulation. For example, দেউলিয়া /d̪eulɾʲɐ/, নিয়তি /nɾʲɔtɪ/, নিয়ম /nʲom/- in these three words the Bangla letter 'য়' is pronounced as palatalized ʲ. দাবায় /d̪ɐbɐɪ̯/, জয় /ɟoɪ̯/ - 'য়' is pronounced as diphthong.

There are a few disputes among linguists regarding Bangla consonants. We have discussed the issues and provided a solution which we have followed in this consonant chart and in the curated dataset.

### 2.4.1. Plosive vs. Affricate Argument

| | চ | ছ | জ | ঝ |
|---|---|---|---|---|
| Plosive | c | cʰ | ɟ | ɟʱ |
| Affricate | tʃ | tʃʰ | dʒ | dʒʱ |

Table 3: Plosive vs Fricative in Bangla

There has been a longstanding dispute among linguists about whether certain Bangla sounds, particularly those represented by চ c, ছ cʰ, জ ɟ, and ঝ ɟʱ, should be classified as affricates or plosives (table 3). Hai (1964) agreed with this discussion and sided with the view that these sounds are best described as palatal plosives. In this proposal, we agree with this perspective, as when we consider how we articulate these words, they seem to align more closely with plosives rather than affricates.

### 2.4.2. ট - Alveolar or Retroflex

| | ট | ঠ |
|---|---|---|
| Alveolar | t | tʰ |
| Retroflex | ʈ | ʈʰ |

Table 4: Alveolar or Retroflex in Bengali

The ট sound in Bangla is produced with the alveolar ridge acting as the fixed point in the mouth (table 4). The active part, which usually includes the tip of the tongue, interacts with this ridge during articulation (Hai, 1964). Abdul Hai (Hai, 1964) acknowledges that while articulating words, the tip of the tongue curls up and back. This is why he categorizes it as an alveolar-retroflex-plosive sound (Hai, 1964).

### 2.4.3. ফ - /pʰ/ and /f/

The pronunciation of the sound represented by ফ in Bangla can vary regionally. While it is generally considered a plosive sound, in some regions, it may be perceived as a labio-dental fricative /f/ (Hai, 1964).

Sometimes native speaker articulates words such as ফরি/fɔɾɪ/, ফাইজলামি/fɐɪɟlɐmɪ/, ফরালেহা/fɔɾɛlɛhɐ/ with a dialectal accent of a certain region. While producing the /pʰ/ sound, they tend to bring the bottom lip close to the upper teeth, creating a narrow passage for the air to flow through. This suggests that ফ can indeed resemble a labio-dental fricative sound /f/. However, it's important to note that this can still be a subject of debate, with variations observed from region to region and from person to person. As for written transcription, without the aid of audio from a regional speaker, accurately determining whether ফ /f/ is pronounced as a plosive or a labio-dental fricative can be challenging. But if we have audio data from regional speakers, we can transcribe words that are pronounced with dialectal accents with /f/ sound (such as fɔralæha) and other words that are also found in the standard Bangla with /pʰ/ (such as pʰul, pʰɔʃol).

Another concern with the /pʰ/ sound is when dealing with borrowed foreign words, there can be further variations in pronunciation. A native speaker of standard Bangla uses the loaned word with a received pronunciation. Hence for the loaned words, the labio-dental f sound has been used for the transcription of the Bangla letter ফ.

#### 2.4.4. Trill r vs. Tap ɾ

The government website employs the trill 'r' sound, but in the Bangla language, for words such as রাজা, রাজ্য, and রাগ we don't naturally produce the trill sound. To ensure better pronunciation, the tap sound (ɾ) would be more suitable for Bangla.

#### 2.4.5. Contextual Substitution of phoneme

The Bangla /ɟ/ is a voiced palatal stop and in standard Bangla, there is no voiced alveolar fricative /z/. Furthermore, in the Bangla language, the closest phoneme with the labio-dental fricatives such as /f/ and /v/ are aspirated labial stops /pʰ/ and /bʱ/. However, many words in standard Bangla are adapted from foreign languages such as English, Arabic, Farsi, and so on. When native speakers articulate these loaned words they do not pronounce them in the same way a native English or native speaker Arabic does, but pronounce these with a native influence. Hence, for loaned words where the speaker articulates these foreign phonemes in a certain word context, we consider these phonemes (/ɟ/, /f/, /v/) in the IPA transcription.

#### 2.4.6. Voiced Aspiration

Aspiration is a distinctive feature in the Bangla phoneme. It can be noted from the chart above that ভ /bʱ/, ধ /dʱ/, ঢ /ɖʱ/, ঝ /ɟʱ/, and ঘ /gʱ/ are voiced aspirated stops. Aspiration is about how much air leaves your mouth while articulating the phoneme. If an unvoiced consonant is aspirated, then an extra puff of air leaves the mouth after the primary articulation is complete. For example in /pʰ/, /tʰ/, /cʰ/, /ʈʰ/, and /kʰ/ voiceless aspiration occurs, hence for the secondary articulation of the aspiration, we use /ʰ/ which is voiceless. On the other hand, /bʱ/, /dʱ/, /ɖʱ/, /ɟʱ/ and /gʱ/ are voiced stops and for that reason, it is suitable to use a voiced aspiration /ʱ/ for the secondary articulation. In the govt-IPA, the aspiration suggestions for voiced stops have both voiced /ʰ/ aspiration and voiceless /ʱ/ aspiration as their secondary articulation. For instance, they kept both /bʰ/ or /bʱ/ for the transcription of the letter 'ভ' despite that the /ʱ/ should be voiced after voiced consonants.

### 2.5. Diacritics

Our proposed diacritics for standard Bangla are /ʷ/ (Labialized), /ʲ/ (Palatalized) and /◌̃/ (Nasalized)

**Labialized:** The use of labialized diacritics is found in Bangla words such as উপরওয়ালা /uporoʷɐlɐ/, দেওয়া /deoʷɐ/, নেওয়া /neoʷɐ/, etc where the consonant sounds indicate that they are pronounced with rounded lips. In certain cases, diphthongs are pronounced with simultaneous lip rounding, such as রওশন /rɔo͡ʷ.ʃon/.

**Palatalized:** To determine the use of palatalized ʲ, we have followed two phonological rules. The rule for determining whether the Bangla consonant য় (j) is palatalized or functions as a diphthong is as follows:

**Case of coda** য়: When the position of the য় is in the syllable-final, without a following vowel, it remains unpalatalized. For example, in compound words like মামলায় /mɐmlɐɪ̯/, নিরাপত্তায় /nirɐpottɐɪ̯/, etc.

**Case of middle** য় Conversely, if a word with য় concludes with a vowel in the syllable's final position and does not have য় in the word's final position, it will be pronounced as a palatalized ʲ. For instance, this can be observed in words like ছেলেমেয়ে /cʰelemeʲe/, খায়রুল /kʰɐʲerul/, and নিয়ম /niʲom/.

**Nasalized:** It was mentioned earlier that in Bangla, all seven oral vowels have their seven nasal counterparts which is described using the nasalized diacritics /ĩ ẽ õ ã ɔ̃ ũ/. This nasalization of vowels in Bangla text is consistently indicated by a diacritic known as 'chandarabindu' (◌̃) placed above the relevant segment, and this occurrence is a common feature in Standard Bangla text.

### 2.6. Loan Words Consideration: Vowel and Consonant

In the Bangla language, using loaned words from foreign languages and using them with a different pronunciation in comparison to their native pronunciation is quite common. In the case of vowels, no foreign phonemes are produced by native speakers. For example, the English word 'foam', 'cloud', and 'flower' is pronounced as /foʊm/, /klaʊd/, and /flaʊə/ by native English speakers. However, /ʊ/ and /ə/ are not articulated by the Bengali native speakers. Instead, they pronounce these words using the existing vowel phonemes of the Bangla language. On the contrary, there are a few cases where foreign words are pronounced using consonant phonemes which does not exist in Bangla.

Labio-dental fricative sounds such as /f/, and /v/ do not exist in the Bangla language but they are articulated by the native speakers when they produce loaned words with these phonemes. In Bengali Some examples are: Plosive ফ (/pʰ/): ফড়িং (/pʰoriŋ/); Plosive ভ (/bʰ/): ভয় (/bʰoe̯/); Fricative (/f/): (ফেইল (/feɪl/)); Fricative (/v/): (ভিউ (/vɪu/))

Same case for the alveolar fricative phoneme /z/. Loaned words from Arabic and English languages such as মেরাজ /merɐz/, ম্যাগাজিন /mɛgɐzɪn/, মোনাজাত /monɐzɐt/ are continuously used in the Standard Bangla. For example, the plosive sound /ɟ/ for জ, য is present in Bengali whereas the Fricative /z/ is found in loan words such as ম্যাগাজিন (/mɛgɐzɪn/) English words such as judge /dʒʌdʒ/, and justice /dʒʌstɪs/ have voiced postalveolar affricate /dʒ/ which is not used by native Bangla speakers. They turn this affricate sound into the plosive sound /ɟ/ and articulate it as /ɟudɟ/ and /ɟustɪs/.

The English language has a voiceless dental fricative sound /θ/ which is not found in the Bangla language. They turn this phoneme into a voiceless aspirated dental plosive sound /t̪ʰ/. So 'think' is pronounced as /t̪ʰiŋk/ in its Bangla adaptive form. The /s/ is a voiceless fricative alveolar sound that is found in both Bangla and other foreign languages such as English.

## 2.7. Validation and Linguistic Challenges of Standard Bengali IPA

### 2.7.1. Morphological Variations in Words

The Bangla language exhibits an extensive array of morphological variations, presenting a challenge in accurately contextualizing the meaning of words in light of their morphological alterations. It poses a challenge to accurately represent these subtle morphological variations within the framework of the International Phonetic Alphabet (IPA).

Consider the Bangla word আজকেই, transcribed as /ɐɟkeː/, or loaned words with Bangla morphological extensions like মেক্সিকোতেও /meksɪkoto:/ and মেক্সিকোও /meksɪkoo:/. While these all end with a vowel, without a syllabic marker, it may not be immediately clear that these suffixes are part of the base word. However, by incorporating the lengthening diacritic after the word (the long vowel diacritic /ː/), this distinction becomes more apparent to the reader.

The reason for utilizing this diacritic is rooted in certain linguistic contexts. In some cases, when producing specific vowels, some individuals perceive a long i: as merely an extended version of the short vowel, without any discernible difference in quality, i.e., without raising the tongue for the long sound. For instance, Bangla e: is slightly higher than Bangla e, and Bangla e̞ (short) falls midway between cardinal e and ɛ. This concept is supported in the work of Suniti Kumar Chatterji as well. Furthermore, this long vowel diacritic also clears out the confusion that no case of diphthongs is present here (মেক্সিকোও /meksɪkoo:/).

The issue with morphological suffixes may create confusion to distinguish them from diphthongs such as the above word গরুগুলোও /goruguloo:/, some might transcribe it as গরুগুলোও /goruguloo͡/ because there are two vowels together in the word. But if we notice carefully and break into the syllable of the /go.ru.gu.lo.o:/, both of the vowels belong to different syllables, even if both of the vowels are beside each other the last vowel o is pronounced with a long sound. This is the reason we have annotated morphological variation in such cases with long vowel marks. Some sample cases are শুটিংয়ে (ʃu.tɪŋʲ.eː), শুটিংও (ʃu.tɪŋ.oː) and গরুগুলোও (goruguloo:).

### 2.7.2. Diphthongs

Our dataset contains cases of Bangla diphthongs. To accurately transcribe them, it's crucial to first identify whether they are indeed diphthongs. Syllabification serves as a method to recognize diphthongs which makes the process easier. However, due to the shortness of time, we decided to avoid the process of syllabication of each word just to identify diphthongs. Another significant aspect in distinguishing diphthongs is the use of the glide. The upper diphthong glide (͡) describes the movement of the articulatory vocal organs, particularly the tongue, from a higher position to a lower one during diphthong production. This downward movement contributes to the distinct sound of the diphthong. Each language possesses its own set of unique diphthongs. We've provided a diphthong chart, from which standard Bangla focuses primarily on the regular diphthongs. Understanding the role of the glide and accurately using it ensures the correct pronunciation of words in a given language.

**Some examples are**
পরিচর্যায় (porɪcɔrɟɐe̯), ভাই (bʰɑi̯), যাচাই (ɟɐ.cɐi̯), চাই (cɐi̯), দুই (dui̯), বোঝাই (bo.ɟʰɑi̯)

Sometimes, a few cases of standard Bangla are found which may cause confusion to the reader, if a certain word has a diphthong or vowel cluster. For example, শিরোইলে is transcribed as /ʃɪroɪle/, here the roɪ constitutes one single syllable, but the question remains if it is a vowel cluster or diphthong. Bangla native speakers articulate this word in this way where a downward movement of tongue position from o to ɪ occurs. As a result, the o stays as a pure vowel and glides toward ɪ which creates a diphthong. Hence, the final transcribed text is /ʃɪro͡ɪle/. If the pronunciation of the word were something such as /ʃɪ.ro.ɪ.le/ where the letters are pronounced as a pure vowel and separately from the syllable then the final result might have been something different.

### 2.7.3. Loan words

Native speakers of the Bangla language commonly integrate vocabulary from English, Arabic, Farsi, and Portuguese into their speech. As a result, distinctive phonemes of these languages, which may not be common in standard Bangla, are spoken by native speakers. Due to their frequent usage, these phonemes may not be distinctly differentiated from the standard Bangla phonetic inventory. This challenges IPA models in accurately recognizing and transcribing these foreign phonetic elements. In our dataset, we have a significant number of English and Arabic words. To transcribe these words, we consider how native Bangla speakers, adhering to the standard Bangla form, would pronounce them. Since standard Bangla users often employ a more received pronunciation when uttering these words, we have annotated them accordingly. Hence, we have used /z/, /f/, /v/, /s/ phonemes for the letters জ/য, ফ, ভ, শ/স respectively. These sounds are not commonly present in the native Bangla language, but to transcribe the borrowed foreign words, we have employed these. **Some examples are**
ফেইক (feɪk), শিডিউল (ʃɪ.dɪ.ul), মোস্তাফিজ (mostɐfɪz), যার-হাদ (zɐrhɐd), ফজর (fɔzor), রড্রিগেজ (rɔdrɪgez)

### 2.7.4. English Diphthong and Triphthong in Bangla Adaptive Form

In English words with diphthongs, the presence of schwa/ə/ can influence the pronunciation. It appears in unstressed syllables, usually containing the neutral, unstressed vowel sound. This leads to subtle variations in how diphthongs are articulated. For example, 'power'- in the word, the diphthong /aʊ/ is followed by the schwa sound in the unstressed syllable. Or for the word 'water', the first syllable may be reduced to a schwa sound, especially if it's unstressed. It might sound like "wuh-ter." However, when these words are adapted by the Bangla speaker they will be pronounced like /pa.o͡ʷɐr/ /o͡ɐ.ter/.

Bangla speakers adopt English diphthongs that do not contain schwa and the pronunciation tends to align with the native English pronunciation. For example, 'high' is transcribed in the Bangla as /h͡ɐɪ/, boil as /b͡ɔɪl/, and time as /t͡ɐɪm/.

The English language contains triphthongs, which is a rare case in the Bangla language. In the case of English triphthongs, native Bangla speakers tend to avoid pronouncing the word as a triphthong. Instead, they convert it into a diphthong and therefore avoid pronouncing the triphthong word. For example, in English, the word 'fire' is pronounced as /fʌɪə/, which in Bangla is transcribed as /fɐʲe̞.ɐr/. Cases like these are found in these words as well - 'hour' /a ər/, which is pronounced as /ɐ.o͡ʷɐr/, 'prayer' /preɪər/, pronounced as /pre.ɐr/, 'pure' /pjʊr/ pronounced as /p͡ɪʊr/.

Hence the only concern while transcribing these words is how a native speaker pronounces them. **Some examples are** ফায়ার (fɐʲe.ɐr), ফাইনাল (f͡ɐɪ.nal), শুটআউটে (ʃut.ɐ͡ʊte:)

In the first example, /fɐʲe.ɐr/ is transcribed for the English word 'fire'. The native English speaker pronounced it as, /faɪər/ where the diphthong /aɪ/ glides into schwa /ə/ in the second syllable. However, the Bangla language does not have a schwa /ə/ sound as a result for this English diphthong word native Bangla speakers use the existing sound to produce the loaned word as /fɐʲe.ɐr/ which does not have a diphthong in the adaptive form.

The pronunciation of words by Bangla speakers can vary based on regional accents and specific contexts. Even a standard native speaker may pronounce certain words differently depending on the situation, which could lead to variations in IPA transcription. Unless the transcription is based on audio data, ensuring accurate contextual transcription can be a challenge.

### 2.7.5. Transcribing Numbers

In the dataset, there are numbers represented in various forms. A combination of letters and numbers ("19টা" 19tɐ, "১ম" 1m) or only a combination of numbers such as "১৯৮৯", "১০০০", or in the context of phone numbers and house numbers, were present. To transcribe these, we followed an IPA transcription based on how we naturally pronounce them. For instance, "২০৬" is transcribed as "d̪uʃo cʰoe̞". When numbers are pronounced individually, they are transcribed accordingly, for example, "২০৫০" as "d̪uɪ ʃunno pãc ʃunno".

### 2.7.6. Handling the cases of Abbreviations and Acronyms

To ensure dataset accuracy and disambiguate between abbreviations and acronyms, we established a specific protocol. When transcribing an abbreviation like "ম."/M/, we consider the context to identify their full forms, which in this case were "মহাম্মদ" /Mɔhɐmmɔd/. We then proceeded to transcribe the entire words. In the case of acronyms like "মূ-সক" /muʃɔk, we applied IPA notation for accurate representation. Handling these types of transcriptions poses certain challenges. Sometimes মহাম্মদ /Mɔhɐmmɔd/ might be spelled and pronounced as মহাম্মাদ /Mɔhɐmmɐd/ or only স. is only given in a sentence and the transcriber has to assume the words if a proper indication is not given in the sentence. So with a large number of acronyms and abbreviations in a language, the transcription of IPA for these may produce incorrect transcriptions. **Some examples are**
এসএসসি (esessɪ), পিডিডি (pɪdɪdɪ), মূসক (muʃɔk)
Some abbreviation examples **are given below in (Abbreviation, Bangla Word (IPA)) template**,
(ম., মহাম্মদ (mɔhɐmmɔd)), (মো., মোহাম্মদ (mohɐmmɔd)), (ডা., ডাক্তার (dɐktɐr))

### 2.7.7. Orthographic Challenges

Bangla orthography may not always align perfectly with phonetic transcription, requiring careful interpretation. Our dataset has been curated from writ-ten texts, based on the specific annotator's pronunciation intuition, as pronunciation sometimes varies from individual to individual. In spite of this, the pronunciation of a word might match word to word in the IPA transcription. Such as হ্রাসমান /rɛʃmɛn/, the হ letter here is not pronounced the way it is pronounced in the word হলুদ /holud̪/. Also in the spelling of the word হলুদ/holud̪/, there is not any 'ও' visible but while articulating the word an /o/ sound has been produced and that's how the word has been transcribed.

### 2.7.8. Placement of Diacritics

IPA transcription involves a meticulous and time-consuming manual process. Accurate placement of diacritics and special characters is critical for correctly representing sounds. For instance, if we were to transcribe the Bengali word দোয়েল as /doel/ or /d̪o͡el/, rather than /d̪oʲel/, it would lead to an inaccurate pronunciation.

## 3. DUAL-IPA Dataset

### 3.1. Dataset Construction

Following the proposed IPA framework, we constructed the DUAL-IPA dataset, containing 150k Bangla sentences along with their linguist-validated IPA transcription. We collected the sentences from two sources: Bangla online newspapers(33%) and literature/books(66%). The sentences have been equally distributed among 4 linguists with a graduate degree in linguistics, along with the above IPA transcription protocol. An independent evaluator has meticulously evaluated all the data to ensure consistency and correctness of annotation. It took a month for the curation of the dataset. The annotation process was expedited using i) **Preannotation:** A rule (and later, a weak model)-based noisy pre-annotation. ii) **Validation:** Word(whitespace separated tokens)-level transcription correction iii) **Mapping** the word level transcription with the sentences. iv) **Sentence level validation** to fix the transcription for fixing the homograph cases, numerals, and alignment errors.

### 3.2. Dataset statistics (EDA)

The dataset contains 150k sentences, with an average of The train split contains 100k sentences and the test split contains 50k sentences. There are about 130k unique words in the training data and 35k out of vocabulary words in the test dataset.

## 4. Benchmarking

We trained a simple LLM-based seq2seq model for benchmarking IPA transcription for Bengali using the proposed Dual-IPA dataset. Here we used the 'small' variant of the **MT5 model** from Google (Xue et al., 2020) for benchmarking. It is a multilingual variant of T5 that was pre-trained on a new Common Crawl-based dataset covering 101 languages. The model was trained for 10 epochs and a 3e-4 learning rate. Our model obtained a WER of 0.1 on the test dataset.

While evaluating the network, we have chosen Word Error Rate(WER) as a metric, to capture the sentence-level overall performance of the IPA transcription network. The obtained high score can be attributed to having a smaller number of homographs and OOV cases where the words from the inferences dataset are familiar to the network.

## 5. Conclusion and Future Work

In this work, we presented a comprehensive study of the IPA standard of Bangla and discussed all the existing points of debate in the literature. We propose a consistent IPA transcription framework for Bangla texts and discuss the nuances in detail. We also present a novel 150k sentence dataset for sequence-to-sequence NLP modeling. This work has the potential to contribute to the field of linguistic theory, NLP dataset creation(the first large-scale sentence-level dataset for Bangla), and also facilitating LLM downstream tasks.